\documentclass[letterpaper]{article} 
\usepackage{aaai2026}  
\usepackage{times}  
\usepackage{helvet}  
\usepackage{courier}  
\usepackage[hyphens]{url}  
\usepackage{graphicx} 
\usepackage{natbib}  
\usepackage{caption} 
\usepackage{algorithm}
\usepackage{algorithmic}
\usepackage{booktabs}
\usepackage{multirow}
\usepackage{amsmath}
\usepackage{amssymb} 

\usepackage{newfloat}
\usepackage{listings}
\DeclareCaptionStyle{ruled}{labelfont=normalfont,labelsep=colon,strut=off} 
\floatstyle{ruled}
\newfloat{listing}{tb}{lst}{}
\floatname{listing}{Listing}
\nocopyright 

\title{Delta-JEPA: Learning Action-Sensitive World Models via Latent Difference Decoding}
\author{
    Zhenghao Zhang\textsuperscript{\rm 1},
    Yuanxiang Wang\textsuperscript{\rm 1}\thanks{Equal contribution as co-second authors.},
    Zhenyu Guan\textsuperscript{\rm 1}\footnotemark[1],
    Yujia Yang\textsuperscript{\rm 1}\footnotemark[1],
    Bingkang Shi\textsuperscript{\rm 2}\footnotemark[1],
    Tianyu Zong\textsuperscript{\rm 1},
    Hongzhu Yi\textsuperscript{\rm 1},
    Guoqing Chao\textsuperscript{\rm 3},
    Xingchen Chen\textsuperscript{\rm 4},
    Tiankun Yang\textsuperscript{\rm 1},
    Chenxi Bao\textsuperscript{\rm 1},
    Tao Yu\textsuperscript{\rm 5},
    Jingjing Zhou\textsuperscript{\rm 1},
    Jungang Xu\textsuperscript{\rm 1}\thanks{Corresponding author.}
}
\affiliations{
    \textsuperscript{\rm 1}School of Computer Science and Technology, University of Chinese Academy of Sciences, Beijing\\
    \textsuperscript{\rm 2}Institute of Information Engineering, Chinese Academy of Sciences, Beijing\\
    \textsuperscript{\rm 3}School of Computer Science and Technology, Harbin Institute of Technology, Weihai\\
    \textsuperscript{\rm 4}Faculty of Computing, Harbin Institute of Technology, Harbin\\
    \textsuperscript{\rm 5}Institute of Automation, Chinese Academy of Sciences, Beijing\\
    zhangzhenghao25@mails.ucas.ac.cn, xujg@ucas.ac.cn
}

\usepackage{bibentry}

\begin{document}

\maketitle

\begin{abstract}
Learning visual world models for planning requires compact latent dynamics that remain sensitive to actions, yet reconstruction-free joint-embedding objectives can collapse to action-insensitive representations. We propose Delta-JEPA, an end-to-end reconstruction-free world model that augments latent forward prediction with a Latent Difference Action Decoder (LDAD). Unlike inverse decoders that infer actions from concatenated endpoint embeddings, LDAD reconstructs the executed action from the latent displacement between consecutive observations. This displacement-level supervision directly regularizes transition geometry: adjacent embeddings cannot collapse without losing action information, and different actions are encouraged to induce distinguishable latent changes for rollout-based planning. Delta-JEPA uses only latent prediction and action reconstruction, avoiding pixel reconstruction and distribution-matching regularizers. Across four visual continuous-control tasks, Delta-JEPA improves planning over JEPA-based and representation-learning world model baselines. Ablations show that displacement-based action decoding is consistently more effective than endpoint concatenation, and action-sensitivity analyses show clearer action-conditioned latent responses. These results indicate that supervising latent differences is a simple and effective mechanism for collapse-resistant and action-sensitive world model learning.
\end{abstract}


\section{Introduction}

Building agents that can infer environment dynamics and predict future states directly from raw sensory observations remains a central goal in artificial intelligence~\citep{ha2018recurrent,ha2018world}. World models address this goal by learning an internal ``imagination space'' in which future outcomes can be forecast under candidate actions, thereby supporting planning and control~\citep{hafner2019dream,wu2023daydreamer}. Early world models often relied on pixel-space reconstruction~\citep{hafner2019learning}, but reconstructing high-dimensional observations is computationally expensive and can waste model capacity on visually detailed but dynamics-irrelevant information~\citep{assran2023self,assran2025v,hauri2026dreamer}. This makes reconstruction-free latent prediction an attractive alternative.

Joint Embedding Predictive Architectures (JEPA)~\citep{assran2023self} offer a particularly appealing foundation for latent world modeling because they directly predict compact future representations rather than future pixels. However, this efficiency introduces a major challenge: when trained end-to-end with only latent prediction objectives, JEPA-based world models can easily collapse to trivial constant representations~\citep{maes2026leworldmodel}. In that case, the model achieves deceptively low prediction loss while destroying the representation structure needed for planning.

Existing approaches typically address collapse through additional training heuristics, though these designs involve different tradeoffs. LeWorldModel~\citep{maes2026leworldmodel}, for example, uses SigReg~\citep{balestriero2025lejepa} to stabilize end-to-end latent prediction, but it does not explicitly constrain the latent space to be sensitive to executed actions, allowing different actions to induce weakly distinguishable latent transitions. PLDM~\citep{sobal2026learning} instead combines VICReg-style regularization with inverse dynamics, yielding a more complex multi-loss objective that is sensitive to hyperparameter tuning. Moreover, its inverse dynamics module decodes actions from concatenated adjacent latent states $[z_t, z_{t+1}]$. Because the forward predictor is itself conditioned on the executed action, end-to-end optimization can make the next-state representation $z_{t+1}$ absorb action-correlated cues that are easy for the inverse decoder to exploit, without requiring the model to represent the actual transition between the two states.

To address these issues, we propose \textbf{Delta-JEPA}, an end-to-end latent world model built around the \textbf{Latent Difference Action Decoder (LDAD)}. Instead of reconstructing actions from concatenated latent states, LDAD predicts the executed action from the latent difference $\Delta z_t = z_{t+1} - z_t$. This displacement-level inverse objective encourages action-sensitive latent dynamics that are crucial for planning: if different actions from the same latent state lead to indistinguishable next embeddings, the world model cannot represent action-controllable next-state transitions, making latent rollouts uninformative for planning. Conversely, a latent representation is more controllable when different actions from the same state induce distinguishable next-state embeddings. By requiring the action to be recovered from $\Delta z_t$, LDAD encourages different actions to induce distinguishable latent displacements and next-state embeddings, while discouraging action prediction from relying on state-specific cues rather than the transition itself.

Delta-JEPA trains this mechanism with a simple two-objective scheme: latent forward prediction models future representations under actions, while LDAD makes action-induced latent displacements predictive of their actions. This design is particularly important for planning, where candidate action sequences are evaluated through latent rollouts and the model must distinguish how alternative actions drive the environment forward. Empirically, we show that Delta-JEPA improves planning performance and learns more action-sensitive latent transition structure across diverse continuous-control tasks.  

The main contributions of this work are summarized as follows:
\begin{itemize}
    \item \textbf{Action-Sensitive Latent Dynamics}: We introduce LDAD, a displacement-based inverse objective that mitigates collapse by enforcing action-distinguishable latent transitions.
    \item \textbf{Two-Objective Training Framework}: We develop Delta-JEPA, an end-to-end latent world model trained only with latent forward prediction and LDAD-based action reconstruction.
    \item \textbf{Empirical Validation}: We evaluate Delta-JEPA on diverse continuous-control tasks and show improved planning performance together with stronger action-sensitive latent dynamics.
\end{itemize}

\section{Related Work}
\subsection{Latent World Models}
World models learn compact predictive models of environment dynamics that support planning and control from high-dimensional observations~\citep{ha2018world,ha2018recurrent}. A prominent line of work builds latent dynamics models for visual control, including PlaNet~\citep{hafner2019learning}, Dreamer~\citep{hafner2019dream}, and DreamerV3~\citep{hafner2023mastering}, which encode pixels into latent states and use imagined rollouts for planning or policy learning. These methods demonstrate the effectiveness of latent imagination, but they commonly rely on reconstruction or reward-driven objectives. This motivates reconstruction-free latent world models that directly predict compact representations and focus model capacity on control-relevant state changes.

\subsection{Joint Embedding Predictive Architectures}
Joint Embedding Predictive Architectures (JEPA) were proposed as non-generative predictive models that compare predictions in representation space rather than input space~\citep{lecun2022path}. I-JEPA instantiates this idea for images by predicting masked target embeddings from context embeddings~\citep{assran2023self}, while V-JEPA extends feature prediction to videos and learns spatiotemporal representations without labels, text supervision, or pixel reconstruction~\citep{bardes2024revisiting}. For world model learning, JEPA is attractive because planning requires accurate predictions of how different actions lead to different future states, rather than photorealistic observation synthesis. However, end-to-end JEPA training with only latent prediction losses can admit trivial constant representations, making collapse prevention a central design issue.

\subsection{Collapse Prevention and Inverse Dynamics}
Recent JEPA-based world models introduce additional constraints to avoid feature collapse. DINO-WM~\citep{zhou2025dino} stabilizes latent dynamics learning by using frozen DINOv2 visual features~\citep{oquab2023dinov2}, but this limits task-specific adaptation of the representation. LeWorldModel trains end-to-end with a SigReg-style Gaussian regularizer to encourage non-collapsed latent features~\citep{maes2026leworldmodel,balestriero2025lejepa}. PLDM combines predictive learning with VICReg-style regularization and inverse dynamics~\citep{sobal2026learning,bardes2021vicreg}, but its action decoder operates on concatenated state embeddings, which can allow action-correlated endpoint cues to support inverse prediction without strongly constraining the transition itself. In contrast, Delta-JEPA applies inverse dynamics directly to latent displacements, using action reconstruction to make action-induced latent differences distinguishable while avoiding frozen encoders and complex multi-term regularization.

\begin{figure*}[t]
\centering
\includegraphics[width=0.8\textwidth]{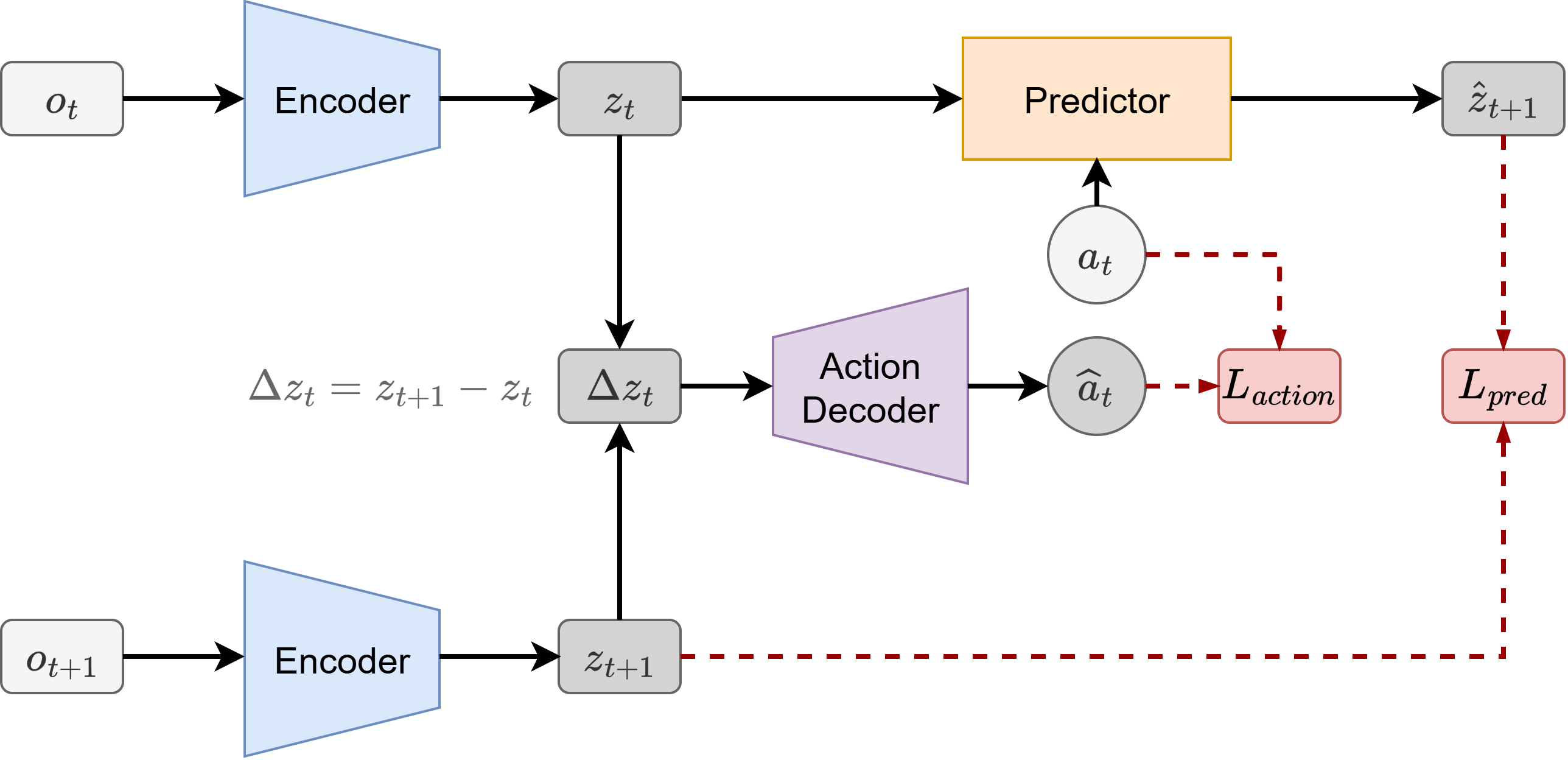} 
\caption{Overview of Delta-JEPA framework. Raw observations $o_t$ and $o_{t+1}$ are mapped to latent representations $z_t$ and $z_{t+1}$ via a shared encoder. In the forward path, the dynamics predictor forecasts the subsequent representation $\hat{z}_{t+1}$ from $z_t$ and the action $a_t$, guided by the prediction loss $\mathcal{L}_{\text{pred}}$. Concurrently, the Latent Difference Action Decoder receives the latent displacement $\Delta z_t$ to reconstruct the action $\hat{a}_t$, supervised by the action loss $\mathcal{L}_{\text{action}}$. This displacement-based action supervision encourages action-induced latent differences to be distinguishable, and the entire framework is optimized end-to-end via $\mathcal{L}= \mathcal{L}_{\text{pred}} + \lambda \mathcal{L}_{\text{action}}$.}
\label{fig:framework}
\end{figure*}
\section{Method}
\subsection{Problem Formulation}
Following the standard paradigm of unsupervised latent world models, we focus on the problem of world model learning in a reward-free, offline setting~\citep{maes2026leworldmodel}. We are given an offline dataset $\mathcal{D} = \{(o_1, a_1, \dots, o_T)\}$ consisting of trajectories with alternating high-dimensional raw image observations $o_t \in \mathbb{R}^{C \times H \times W}$ and continuous actions $a_t \in \mathbb{R}^{d_a}$. Crucially, $\mathcal{D}$ contains no task-specific reward signals and is collected by arbitrary, unknown behavior policies. 

Our goal is to learn a compact latent representation space $\mathcal{Z} \subseteq \mathbb{R}^d$ with an action-sensitive latent dynamics predictor, without reconstructing pixels or using task rewards.

\subsection{Overview of Delta-JEPA}
As illustrated in Figure~\ref{fig:framework}, Delta-JEPA consists of two coupled objectives. The latent forward dynamics predictor learns to forecast the next representation from the current representation and action, providing the rollout model required for planning. The Latent Difference Action Decoder (LDAD) adds an inverse-dynamics constraint on the displacement between adjacent latent states, requiring this displacement to recover the action that caused the transition. Together, these objectives train an end-to-end reconstruction-free world model that discourages collapse to action-insensitive representations and promotes action-sensitive next-state predictions.

\subsection{Latent Forward Dynamics Predictor}
The encoder $f_\theta$ maps each observation $o_t$ to a latent representation $z_t = f_\theta(o_t)$. Conditioned on $z_t$ and action $a_t$, the dynamics predictor $P_\phi$ estimates the next latent state:
\begin{equation}
\label{eq:predictor}
    \hat{z}_{t+1} = P_\phi(z_t, a_t),
\end{equation}
where $\hat{z}_{t+1}$ represents the predicted next latent state.

We train the encoder and predictor with a mean-squared prediction loss in latent space:
\begin{equation}
\label{eq:L_pred}
    \mathcal{L}_{\text{pred}} = \left\| \hat{z}_{t+1} - z_{t+1} \right\|_2^2,
\end{equation}
where $z_{t+1} = f_\theta(o_{t+1})$ is the target representation produced by the same encoder.

Although Eq.~\eqref{eq:L_pred} enables reconstruction-free dynamics learning, it is degenerate when used alone: the encoder and predictor can reduce the loss by collapsing to nearly constant representations. Such a solution preserves little information for planning even if the prediction loss is small. LDAD addresses this failure mode by adding an action-grounded constraint on the difference between adjacent latent states.

\subsection{Latent Difference Action Decoder (LDAD)}
\label{subsec:ldad}

LDAD imposes an inverse-dynamics constraint on the difference between adjacent latent states. Given two encoded observations $z_t$ and $z_{t+1}$, we define the latent displacement as
\begin{equation}
    \Delta z_t = z_{t+1} - z_t .
\end{equation}
The decoder then predicts the executed action from this displacement:
\begin{equation}
\label{eq:decoder}
    \hat{a}_t = D_\Theta(\Delta z_t),
\end{equation}
where $D_\Theta$ denotes the action decoder and $\hat{a}_t$ denotes the predicted action. The decoder is trained end-to-end with a mean-squared action reconstruction loss:
\begin{equation}
\label{eq:L_action}
    \mathcal{L}_{\text{action}} = \left\| \hat{a}_{t} - a_{t} \right\|_2^2.
\end{equation}

\begin{figure}[t]
\centering
\includegraphics[width=0.95\columnwidth]{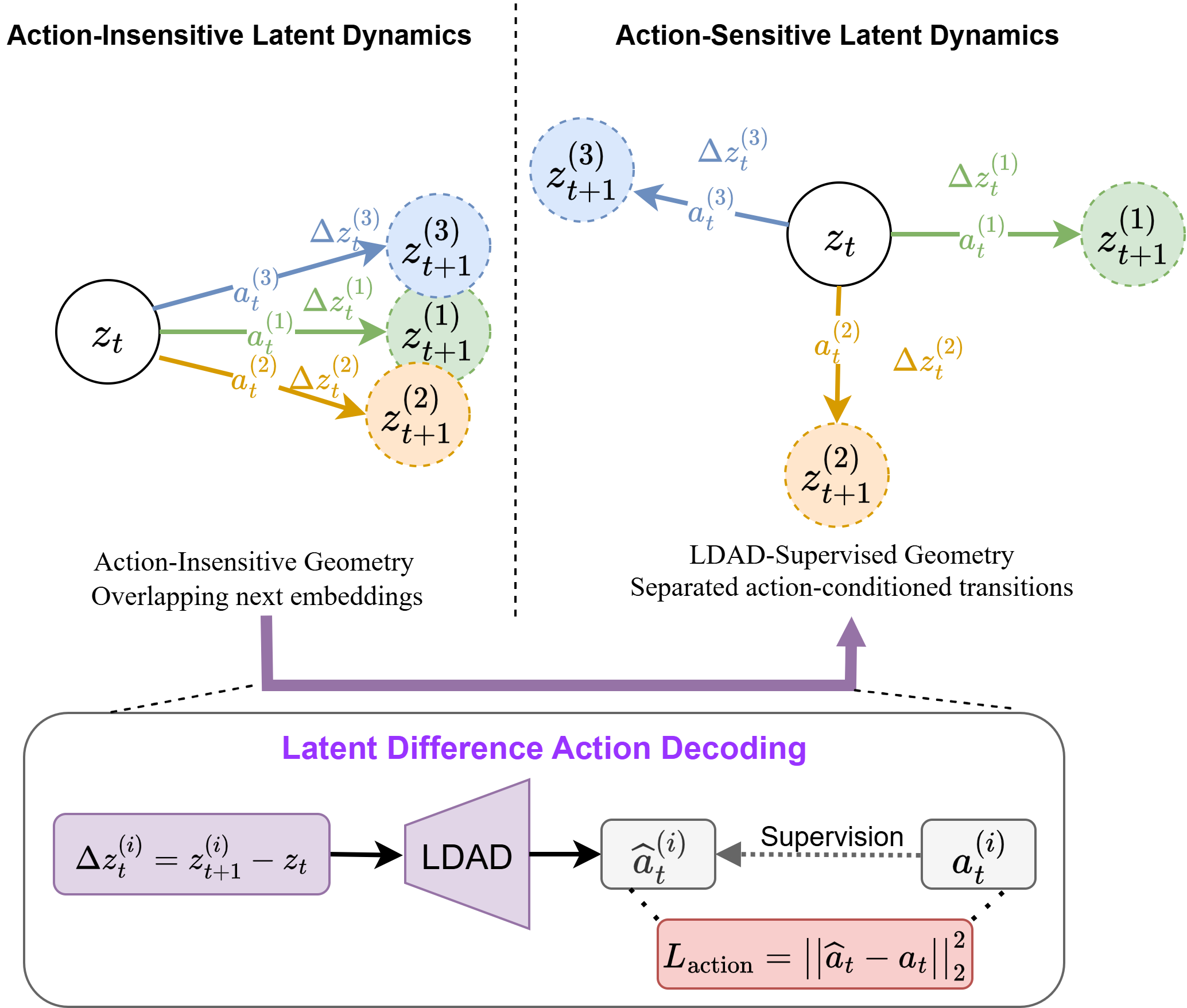}
\caption{Illustration of LDAD-induced action-sensitive latent geometry. Without displacement-level action supervision (top left), different actions from the same latent state $z_t$ may produce similar next embeddings. LDAD computes each displacement $\Delta z_t^{(i)} = z_{t+1}^{(i)} - z_t$, decodes the action $\hat{a}_t^{(i)}$, and supervises it with $\mathcal{L}_{\text{action}} = \|\hat{a}_t-a_t\|_2^2$ (bottom). This encourages action-conditioned transitions to occupy distinguishable directions and endpoints in latent space (top right).}
\label{fig:ldad_planning}
\end{figure}

\subsubsection{Action-Supervised Displacement Mechanism.}
As illustrated in Figure~\ref{fig:ldad_planning}, the top-left panel shows an action-insensitive latent geometry: different actions from the same $z_t$ can produce nearby next embeddings, making the latent transition difficult to distinguish by action. LDAD addresses this failure mode through the decoding pipeline shown at the bottom. For each transition, it computes the displacement $\Delta z_t^{(i)} = z_{t+1}^{(i)}-z_t$, predicts the corresponding action $\hat{a}_t^{(i)}$, and optimizes the reconstruction loss against the executed action $a_t^{(i)}$. Since the decoder observes only the displacement, successful action recovery requires the local transition geometry to encode the executed action, thereby encouraging action-induced displacements to become distinguishable.

The top-right panel depicts the intended effect of this supervision: different actions induce separated transition directions and next embeddings. This geometry is particularly important for planning. When different candidate actions lead to similar latent endpoints, rollouts provide little evidence for comparing their consequences and can therefore cause the planner to select ambiguous or incorrect actions. By contrast, separated action-conditioned transitions make candidate rollouts more action-controllable and more informative for planning. The Two-Room trajectory visualization in Figure~\ref{fig:pca_trace_tworoom} provides empirical evidence consistent with this mechanism, showing trajectories with nearby initial states progressively separating under Delta-JEPA as their action-conditioned rollouts diverge. Complementarily, the action-response PCA in Figure~\ref{fig:mean_delta_pca} directly probes the learned predictor by fixing the starting history and varying only the action input, showing that Delta-JEPA produces clearly separated action-wise responses whereas LeWM remains concentrated near the origin.

\subsubsection{Effects of Displacement-Based Action Decoding.}
The displacement-based inverse objective affects the learned representation in three ways:
\begin{enumerate}
    \item \textbf{Anti-Collapse Effect.} The action reconstruction objective discourages the encoder from mapping consecutive observations to nearly identical latent vectors. If adjacent observations collapse, then $\Delta z_t$ becomes uninformative and $D_\Theta$ cannot recover the executed action.
    \item \textbf{Reducing Dependence on Single-State Cues.} A standard inverse dynamics decoder predicts actions from concatenated latent states, $\hat{a}_t = D_\Theta([z_t,z_{t+1}])$. In our setting, this formulation can admit shortcuts: because the forward predictor receives $a_t$ when predicting $\hat{z}_{t+1}$, the learned target representation $z_{t+1}$ may contain action-correlated cues that allow the inverse decoder to recover $a_t$ without strongly modeling the transition itself. LDAD reduces this risk by conditioning the decoder only on the relative displacement $\Delta z_t$, so action reconstruction must be supported by the change between adjacent latent states rather than by state-specific cues.
    \item \textbf{Action-Sensitive Latent Dynamics for Planning.} For planning, the latent representation must support action-conditioned latent rollouts. LDAD encourages different actions from the same latent state to produce distinguishable latent displacements and next-state embeddings. As a result, candidate actions can be compared through the distinct latent rollouts they induce, providing more informative predictions for action selection.
\end{enumerate}

\subsection{Multi-Step Action Decoding}
We implement $D_\Theta$ with a Transformer backbone and extend LDAD to multi-step action decoding to capture longer-horizon temporal structure. Given a horizon $N \ge 1$, the decoder reconstructs the sequence of actions spanning the interval from $t$ to $t+N$ using the long-horizon latent displacement:
\begin{equation}
    \{\hat{a}_\tau\}_{\tau=t}^{t+N-1} = D_\Theta(z_{t+N} - z_t).
\end{equation}
The multi-step LDAD action decoder uses a Transformer with $N$ learnable action queries. The displacement $z_{t+N} - z_t$ is injected into each query through Adaptive Layer Normalization (AdaLN), after which the Transformer layers produce the $N$ reconstructed continuous actions. This multi-step extension imposes an action-grounded dynamics constraint over longer temporal intervals in latent space.

\subsection{Joint Optimization and End-to-End Training}
Ultimately, the overall training objective of our framework is formulated as a joint loss comprising the forward prediction loss and the action reconstruction loss:
\begin{equation}
    \mathcal{L} = \mathcal{L}_{\text{pred}} + \lambda \mathcal{L}_{\text{action}},
\end{equation}
where $\lambda > 0$ is a balancing hyperparameter.

Delta-JEPA uses only two objectives: latent prediction learns action-conditioned dynamics, and action reconstruction makes local latent transitions action-sensitive. It requires no frozen encoders, stop-gradient branches, or distribution-matching regularizers.

\begin{table*}[t]
\centering
\caption{Planning success rate (\%, higher is better) on four continuous-control environments. Bold numbers indicate the best performance in each environment.}
\label{tab:planning_success_rate}
\vspace{0.5em}
\begin{tabular}{lcccc}
\toprule
\textbf{Method} & \textbf{Two-Room} & \textbf{Reacher} & \textbf{Push-T} & \textbf{OGB-Cube} \\
\midrule
PLDM                     & $93.73_{\pm1.03}$ & $64.33_{\pm2.14}$ & $76.13_{\pm1.70}$ & $57.27_{\pm1.53}$ \\
LeWM                     & $74.93_{\pm0.42}$ & $79.87_{\pm0.90}$ & $84.53_{\pm1.50}$ & $64.13_{\pm1.89}$ \\
Sub-JEPA                 & $90.60_{\pm0.53}$ & $81.00_{\pm2.40}$ & $63.73_{\pm0.12}$ & $62.67_{\pm1.45}$ \\
\midrule
\textbf{Delta-JEPA (Ours)}& $\mathbf{100.00_{\pm 0.00}}$ & $\mathbf{81.33_{\pm 0.50}}$ & $\mathbf{89.07_{\pm1.90}}$ & $\mathbf{79.27_{\pm 1.81}}$ \\
\bottomrule
\end{tabular}
\end{table*}

\section{Experiments}

\subsection{Experimental Setup}
\label{subsec:planning_experiments}

\textbf{Environments.} We evaluate Delta-JEPA on four diverse continuous-control tasks:
\begin{itemize}
    \item \textit{Push-T}~\citep{chi2025diffusion}: A 2D non-prehensile manipulation task in which the agent pushes a T-shaped object to a target pose through physical contact.
    \item \textit{Reacher}~\citep{tassa2018deepmind}: A continuous-control task in which the agent controls a two-link planar robotic arm to reach a randomly spawned target.
    \item \textit{Cube}~\citep{park2025ogbench}: A 3D robotic manipulation task in which the agent controls a gripper to relocate a cube to a target 3D position.
    \item \textit{Two-Room}~\citep{zhou2025dino}: A 2D continuous-navigation task in which the agent navigates through a two-room maze to a designated target point.
\end{itemize}

\textbf{Baselines.} We compare Delta-JEPA with several state-of-the-art JEPA-based and representation-learning world models:
\begin{itemize}
    \item \textbf{LeWorldModel (LeWM)}~\citep{maes2026leworldmodel}: Our primary baseline and foundation, which combines next-step latent representation prediction with Gaussian latent-space regularization to enable stable end-to-end JEPA training directly from raw pixels.
    \item \textbf{Sub-JEPA}~\citep{zhao2026sub}: An extension of LeWM that introduces subspace Gaussian regularization to further improve training stability and representation quality.
    \item \textbf{PLDM}~\citep{sobal2026learning}: An end-to-end pixel-based world model that relies on a compound objective comprising VICReg, inverse dynamics, and temporal smoothness terms, making hyperparameter tuning highly cumbersome.
\end{itemize}

\textbf{Implementation Details.} To ensure a fair comparison, we keep the evaluation protocol and the network architectures of our encoder and predictor consistent with those of LeWM. Specifically, the visual encoder $f_\theta$ is instantiated as a randomly initialized ViT-Tiny. The dynamics predictor is parameterized as a 6-layer causal Transformer (16 attention heads, a head dimension of 64, and an MLP hidden dimension of 2048), where action-conditioning features are injected through Adaptive Layer Normalization for state prediction. 
To minimize the computational overhead of action decoding, we implement the action decoder as a lightweight 3-layer non-causal Transformer with $N=5$ learnable action queries, 8 attention heads, a head dimension of 64, and an FFN hidden dimension of 512.

\subsection{Planning Performance}
We first report planning success rates under an evaluation protocol consistent with LeWM. During training and evaluation, Delta-JEPA, PLDM, Sub-JEPA, and LeWM are trained from scratch for 50 epochs. Specifically, Delta-JEPA is optimized with a learning rate of $5 \times 10^{-5}$, and the action reconstruction weight $\lambda$ is set to $10.0$. For PLDM, Sub-JEPA, and LeWM, we follow their respective official training configurations. We randomly sample 50 and 500 trajectories from each environment to construct the validation and test sets, respectively. All methods reported in Table~\ref{tab:planning_success_rate} are independently evaluated over 3 random seeds. The mean planning success rates on the test set are summarized in Table~\ref{tab:planning_success_rate}.

Delta-JEPA achieves the highest mean planning success rate across all four environments. The improvement is most pronounced on OGB-Cube, where Delta-JEPA exceeds the strongest baseline by 15.14 percentage points, and on Two-Room, where it improves over PLDM by 6.27 points. On Push-T, Delta-JEPA improves over LeWM by 4.54 points, indicating that LDAD benefits contact-rich manipulation. On Reacher, where Sub-JEPA already performs strongly, Delta-JEPA still obtains the best mean result with a smaller margin. Overall, these results suggest that action reconstruction from latent displacements helps the predictor distinguish action-dependent outcomes, leading to stronger planning performance across navigation and manipulation tasks.

\subsection{Ablation Study}
\subsubsection{Action Reconstruction Weight.}
\begin{figure}[t]
\centering
\includegraphics[width=\linewidth]{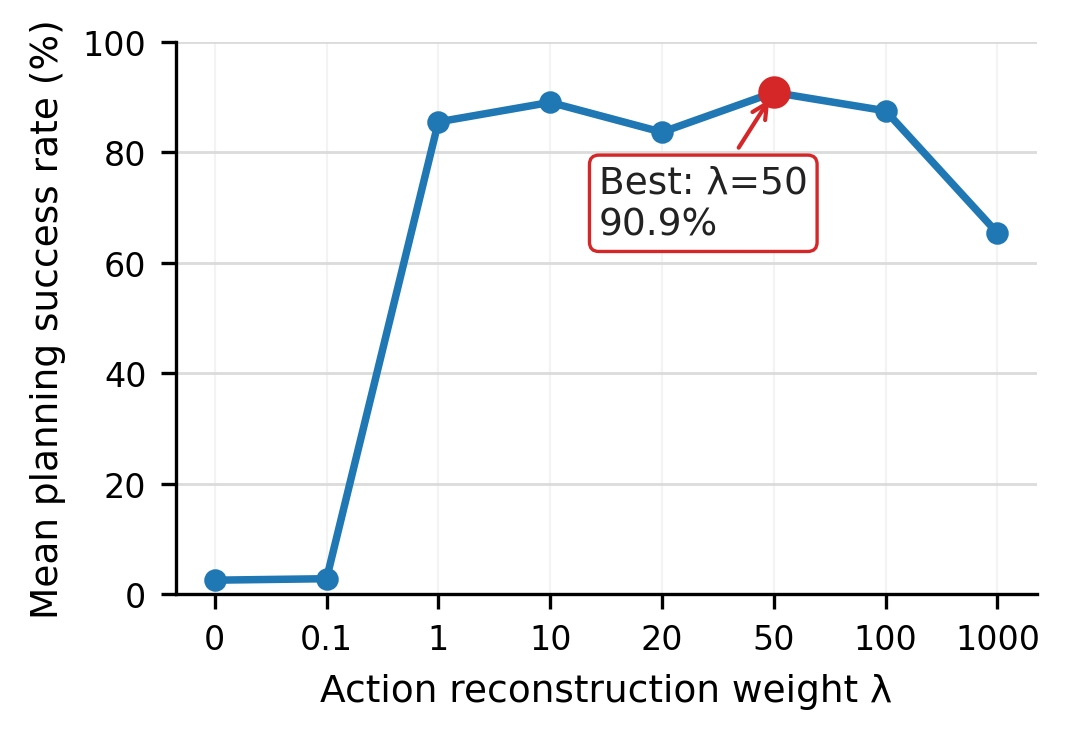}
\caption{Sensitivity of Push-T planning success to the action reconstruction weight $\lambda$. The curve reports the mean success rate over 3 runs, and the peak performance is highlighted.}
\label{fig:action_weight_ablation}
\end{figure}

To evaluate the impact of the proposed LDAD, we conduct a sensitivity analysis of the action reconstruction weight $\lambda$ in the Push-T environment. Specifically, we vary $\lambda$ over the candidate set $\{0, 0.1, 1.0, 10.0, 20.0, 50.0, 100.0, 1000.0\}$. As shown in Figure~\ref{fig:action_weight_ablation}, setting $\lambda=0$ removes LDAD entirely, and the resulting model nearly collapses, yielding only a negligible planning success rate. When $\lambda=0.1$, the LDAD signal remains too weak to provide effective regularization, and the model still performs poorly. In contrast, once $\lambda$ falls within a reasonable range, the planning performance becomes substantially higher and remains relatively stable, with the best result obtained at $\lambda=50.0$. Performance degrades again when the action reconstruction weight is excessively large.

\subsubsection{Displacement-Based Action Decoding.}
To evaluate whether displacement-based action decoding improves downstream planning, we compare LDAD with a variant that reconstructs actions from the concatenated endpoint embeddings $[z_t,z_{t+1}]$ instead of the displacement $\Delta z_t=z_{t+1}-z_t$. Both variants use the same training and evaluation protocol and differ only in the action-decoder input, allowing us to isolate how this design choice affects planning success.

\begin{table*}[t]
\centering
\caption{Ablation of the action-decoder input representation. The concat variant decodes actions from $[z_t,z_{t+1}]$, whereas LDAD decodes actions from $\Delta z_t=z_{t+1}-z_t$. Values are planning success rates (\%) over three seeds.}
\label{tab:z_concat_ablation}
\begin{tabular}{lcccc}
\toprule
\textbf{Action-Decoder Input} & \textbf{Two-Room} & \textbf{Reacher} & \textbf{Push-T} & \textbf{OGB-Cube} \\
\midrule
$[z_t,z_{t+1}]$ & $95.93\pm0.61$ & $80.27\pm0.81$ & $76.47\pm2.08$ & $78.60\pm3.29$ \\
\textbf{$\Delta z_t$ (LDAD)} & $\mathbf{100.00\pm0.00}$ & $\mathbf{81.33\pm0.50}$ & $\mathbf{89.07\pm1.90}$ & $\mathbf{79.27\pm1.81}$ \\
\midrule
Gain & $+4.07$ & $+1.07$ & $+12.60$ & $+0.67$ \\
\bottomrule
\end{tabular}
\end{table*}

As shown in Table~\ref{tab:z_concat_ablation}, using $\Delta z_t$ as the action-decoder input improves planning success on all four environments. The gain is largest on Push-T ($+12.60$ points), followed by Two-Room ($+4.07$ points), while Reacher and OGB-Cube show smaller but consistent improvements. These results indicate that, under the same planning protocol, reconstructing actions from latent displacements provides a more effective training signal for action-conditioned rollouts than reconstructing actions from concatenated endpoint embeddings.

\subsubsection{LDAD Decoding Target.}
\begin{table}[t]
\centering
\caption{Ablation of LDAD decoding targets on Reacher.}
\label{tab:action_target_ablation_reacher}
\begin{tabular}{lc}
\toprule
\textbf{Decoded Target} & \textbf{Planning Success Rate (\%)} \\
\midrule
Raw action $a_t$ & $81.33\pm0.50$ \\
$\Delta$ finger position & $64.93\pm1.10$ \\
$\Delta$ joint position & $80.47\pm2.10$ \\
$\Delta$ finger and joint position & $76.40\pm1.40$ \\
\bottomrule
\end{tabular}
\end{table}

We further ablate the decoding target used by LDAD on Reacher. Besides the raw action $a_t$, we replace the action reconstruction target with state-delta proxies derived from the agent state, including $\Delta$ finger position, $\Delta$ joint position, and their concatenation. As shown in Table~\ref{tab:action_target_ablation_reacher}, decoding raw actions performs best, while $\Delta$ joint position achieves comparable performance and substantially outperforms $\Delta$ finger position. This suggests that LDAD benefits from targets that are tightly aligned with the controllable transition structure of the agent. Notably, concatenating $\Delta$ finger position with $\Delta$ joint position does not further improve performance, indicating that adding extra state-change signals may introduce redundant or less action-aligned information rather than strengthening the displacement supervision.

\subsection{Latent Diversity and Collapse Prevention}
\begin{figure}[t]
\centering
\includegraphics[width=\linewidth]{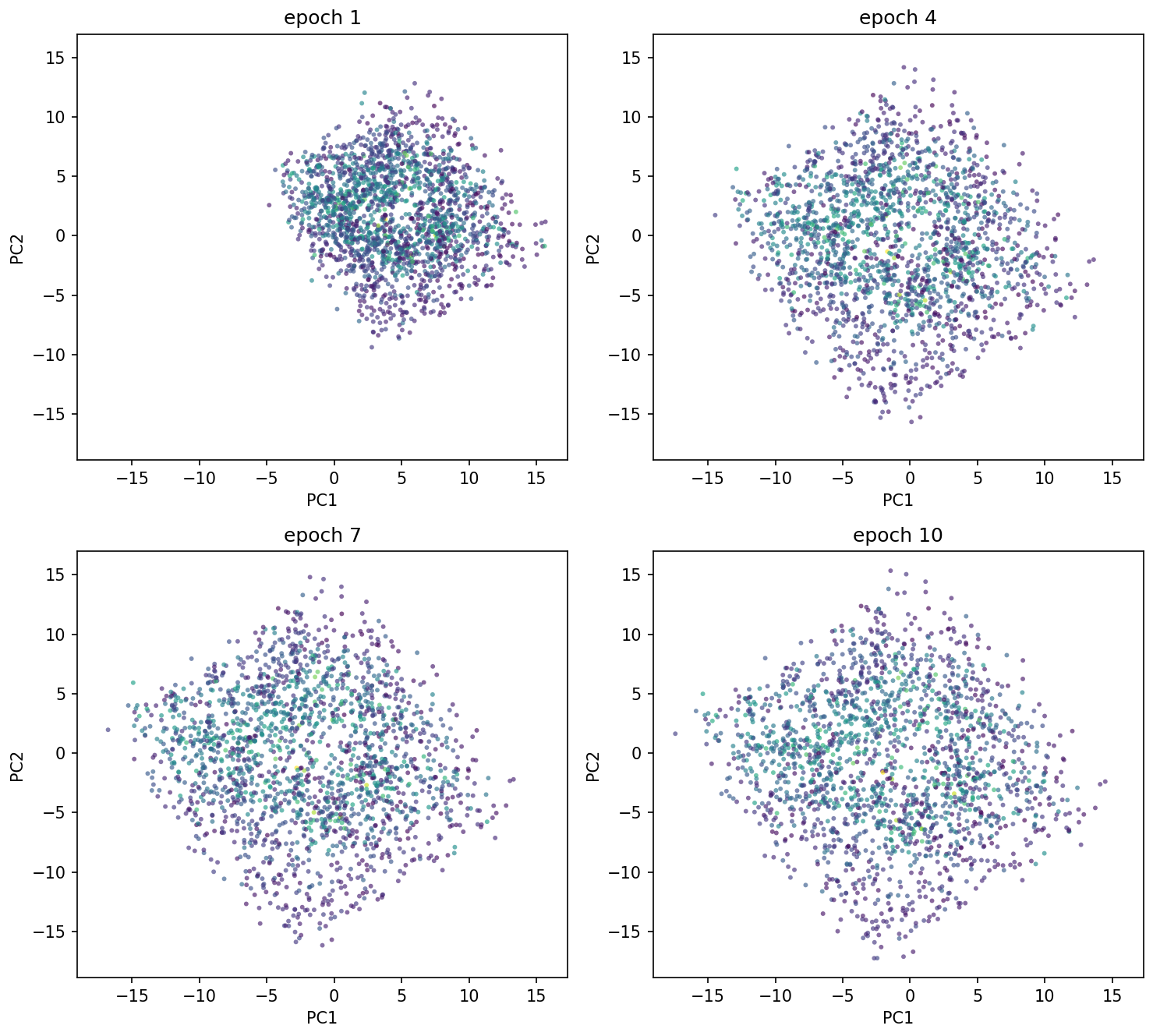}
\caption{Evolution of the learned latent space on Push-T visualized by PCA on 2000 latent representations.}
\label{fig:pca_visualization}
\end{figure}

To qualitatively examine the structure of the learned latent space on Push-T, we apply Principal Component Analysis (PCA) \citep{abdi2010principal} to 2000 latent representations extracted by the encoder. Figure~\ref{fig:pca_visualization} presents the resulting projections at epochs 1, 4, 7, and 10. In the early stage of training, the representations are concentrated within a relatively compact region, suggesting limited latent diversity. As training progresses, they gradually expand over a broader region and form more discernible structures. This trend indicates that Delta-JEPA mitigates representation collapse and learns increasingly discriminative features.

\subsection{Action-Sensitive Latent Dynamics}

\begin{figure*}[t]
\centering
\includegraphics[width=0.9\textwidth]{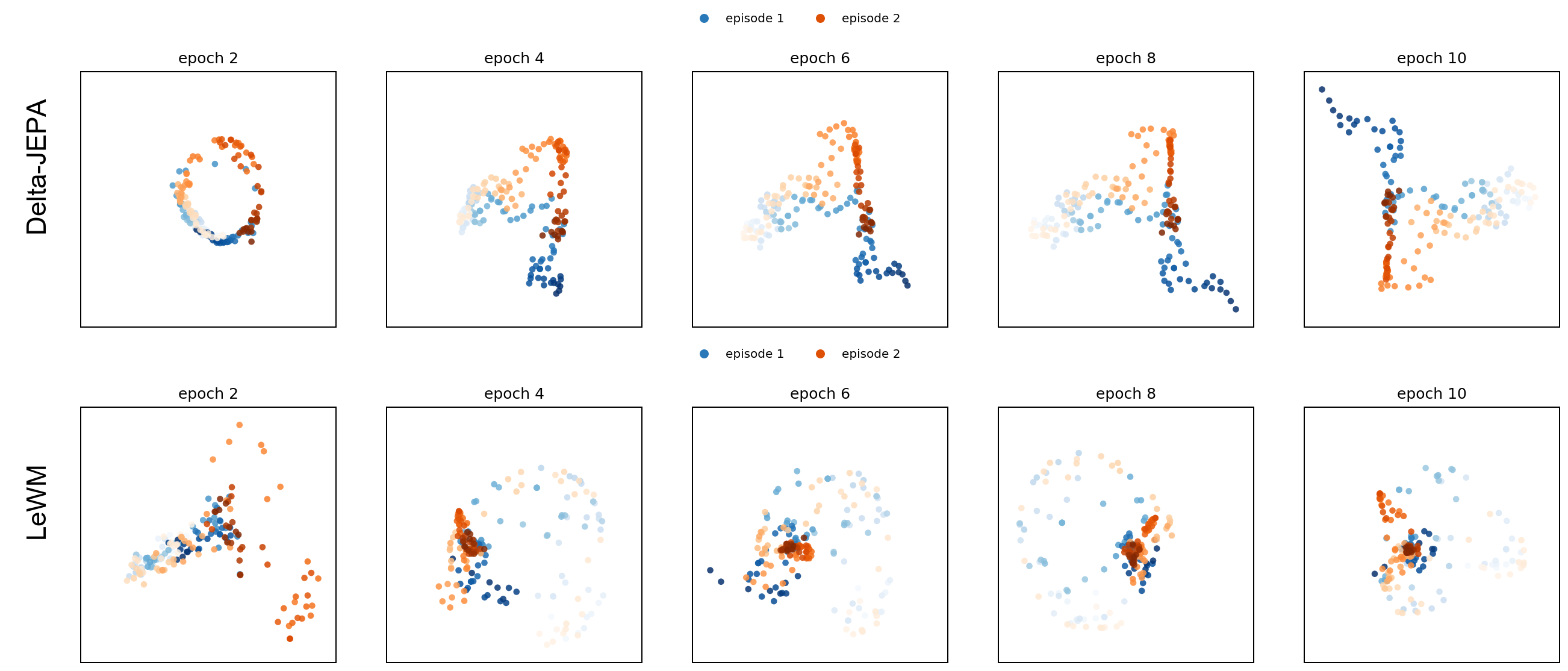}
\caption{PCA visualization of two Two-Room latent trajectories with nearby initial states but different endpoints, shown across training epochs for Delta-JEPA (top) and LeWM (bottom). Blue and orange denote the two trajectories, and color intensity indicates temporal progression from early states (light) to later states (dark).}
\label{fig:pca_trace_tworoom}
\end{figure*}

We further compare Delta-JEPA and LeWM on two Two-Room trajectories selected to have nearby initial states but different endpoints, as shown in Figure~\ref{fig:pca_trace_tworoom}. Each point denotes a latent representation; blue and orange indicate the two trajectories, and darker colors correspond to later timesteps. Delta-JEPA exhibits clear temporal compositionality: the two trajectories start close in latent space and gradually separate as their action-conditioned rollouts diverge. This behavior is consistent with the LDAD mechanism, which encourages latent displacements to preserve action-dependent transition information. LeWM, by contrast, separates some features but produces a less organized geometry, with trajectories that are more scattered and less clearly aligned with temporal progression or action-controllable rollout structure.

\begin{figure}[t]
\centering
\includegraphics[width=0.5\textwidth]{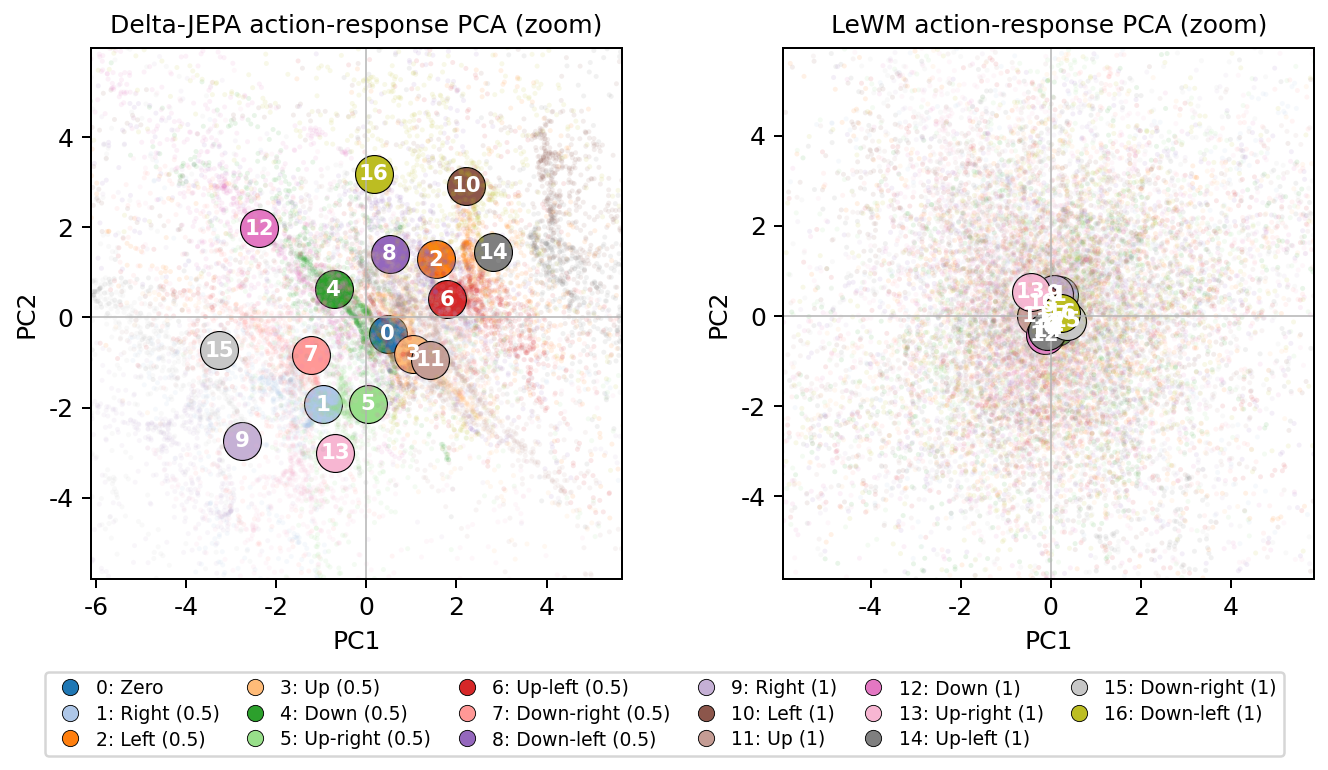}
\caption{PCA visualization of action-conditioned predictor responses on Two-Room. We sample 512 starting histories and keep the history representation $z_t$ fixed while replacing the final action with each candidate action. For each candidate action, we visualize the predicted displacement relative to the zero-action prediction. Each translucent point corresponds to one starting history under one candidate action, and each numbered marker shows the mean response of that candidate action across all 512 histories.}
\label{fig:mean_delta_pca}
\end{figure}

To directly test whether the predictor responds consistently to action changes, we sample 512 Two-Room starting histories and keep each history representation fixed while varying only the final action input. For each candidate action $a$, we compute the predicted next representation $\hat{z}_{t+1}(a)$ and measure its displacement relative to the zero-action prediction, $\hat{z}_{t+1}(a)-\hat{z}_{t+1}(0)$. Figure~\ref{fig:mean_delta_pca} shows a zoomed view centered on the zero-action response, making the action-wise mean markers easier to distinguish. Delta-JEPA produces well-separated action-wise mean responses, with larger action magnitudes generally inducing larger predicted shifts. In contrast, LeWM's action-wise means remain concentrated near the origin and substantially overlap, indicating that changing the action does not induce a stable directional change in its prediction. These results show that Delta-JEPA learns predictor dynamics that are more consistently conditioned on the action input.

\begin{figure*}[!t]
\centering
\includegraphics[width=0.9\textwidth]{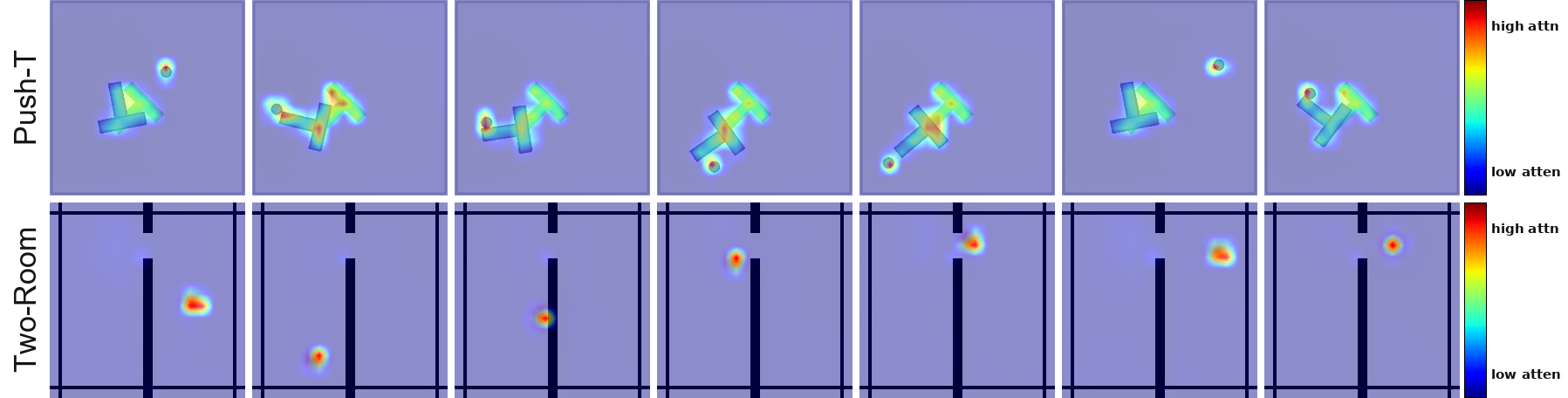}
\caption{Attention rollout visualizations on Push-T (top) and Two-Room (bottom) 
using intermediate layers 4--6 of the ViT-Tiny encoder. Warmer colors indicate 
higher attention weights.}
\label{fig:attention_maps}
\end{figure*}

\subsection{Physical and State-Delta Probing}
To evaluate whether the learned representations preserve underlying environment information, we freeze the encoder and train linear and multi-layer perceptron probes to decode task-specific ground-truth physical attributes from latent states, including agent, object, and end-effector states. For each task, we sample 20,000 observations and split train/test data by trajectory. Table~\ref{tab:physical_probe_tworoom} reports the Two-Room results, and the full probe results on the remaining environments are provided in Appendix~\ref{app:physical_state_probing}. Lower MSE and higher $r$ indicate better representational quality.


\begin{table}[t]
\centering
\caption{Physical latent probing results on Two-Room. Lower MSE and higher $r$ indicate better representational quality.}
\label{tab:physical_probe_tworoom}
\resizebox{\columnwidth}{!}{
\begin{tabular}{llcccc}
\toprule
& & \multicolumn{2}{c}{\textbf{Linear}} & \multicolumn{2}{c}{\textbf{MLP}} \\
\cmidrule(lr){3-4} \cmidrule(lr){5-6}
\textbf{Property} & \textbf{Method} & \textbf{MSE $\downarrow$} & \textbf{$r \uparrow$} & \textbf{MSE $\downarrow$} & \textbf{$r \uparrow$} \\
\midrule
\multirow{5}{*}{Agent Pos.}
& PLDM     & $0.078$ & $0.960$ & $0.002$ & $0.999$ \\
& Sub-JEPA & $0.179$ & $0.907$ & $0.006$ & $0.997$ \\
& LeWM     & $0.085$ & $0.950$ & $0.002$ & $0.999$ \\
& \textbf{Delta-JEPA} & $0.004$ & $0.998$ & $0.000$ & $1.000$ \\
\bottomrule
\end{tabular}
}
\end{table}

We use the same probing protocol as above to evaluate whether latent displacements encode environment changes. Specifically, instead of decoding physical attributes $x_t$ from a single latent state $z_t$, we train probes to predict state changes $\Delta x_t=x_{t+1}-x_t$ from latent displacements $\Delta z_t=z_{t+1}-z_t$. For each task, we sample 20,000 consecutive timestep pairs, split train/test data by trajectory, and train both linear and MLP probes over three random seeds. Table~\ref{tab:state_delta_probe_tworoom} reports the Two-Room results, and Appendix~\ref{app:state_delta_probing} provides the corresponding results on Push-T, DMC Reacher, and OGB-Cube. Lower MSE and higher $r$ indicate that latent displacements better preserve the direction and magnitude of the corresponding physical or task-state changes.

\begin{table}[t]
\centering
\caption{State-delta probing results on Two-Room. The probe predicts $\Delta x_t=x_{t+1}-x_t$ from $\Delta z_t=z_{t+1}-z_t$. Lower MSE and higher $r$ indicate better alignment between latent displacements and physical state changes.}
\label{tab:state_delta_probe_tworoom}
\resizebox{\columnwidth}{!}{
\begin{tabular}{llcccc}
\toprule
& & \multicolumn{2}{c}{\textbf{Linear}} & \multicolumn{2}{c}{\textbf{MLP}} \\
\cmidrule(lr){3-4} \cmidrule(lr){5-6}
\textbf{Property} & \textbf{Method} & \textbf{MSE $\downarrow$} & \textbf{$r \uparrow$} & \textbf{MSE $\downarrow$} & \textbf{$r \uparrow$} \\
\midrule
\multirow{5}{*}{$\Delta$ Agent Pos.}
& PLDM     & $0.355$ & $0.813$ & $0.095$ & $0.955$ \\
& Sub-JEPA & $0.601$ & $0.674$ & $0.141$ & $0.928$ \\
& LeWM     & $0.444$ & $0.765$ & $0.085$ & $0.958$ \\
& \textbf{Delta-JEPA} & $0.016$ & $0.992$ & $0.005$ & $0.997$ \\
\bottomrule
\end{tabular}
}
\end{table}

\subsection{Task-Relevant Attention Patterns} 

To further assess the interpretability of the learned latent representations, we visualize the self-attention patterns of the ViT-Tiny encoder on the Push-T and Two-Room tasks. We employ attention rollout on intermediate transformer blocks and report heatmaps from layers 4--6, where object-related cues are expected to be captured before being integrated into higher-level task representations. As shown in Figure~\ref{fig:attention_maps}, although the model is trained without dense pixel-level reconstruction or explicit object-level supervision, the attention maps concentrate on task-relevant regions, including the agent and the T-shaped block, while assigning relatively low attention to background areas. Additional layer-wise attention visualizations in Appendix~\ref{app:attention_specialization} further show that different encoder layers can emphasize different task-relevant entities. Together, these qualitative results suggest that Delta-JEPA learns compact representations that preserve physically meaningful and object-centric visual structure across environments.

\section{Conclusion}
We proposed Delta-JEPA, a reconstruction-free latent world model that uses Latent Difference Action Decoding to supervise action information directly in latent displacements. By reconstructing actions from $\Delta z_t=z_{t+1}-z_t$, Delta-JEPA discourages collapse and encourages different actions to induce distinguishable latent transitions for planning, while retaining a simple objective that combines latent prediction with action reconstruction. Experiments across four continuous-control tasks show that Delta-JEPA improves planning performance over JEPA-based and representation-learning baselines, and ablations confirm the advantage of displacement-based decoding over endpoint concatenation. Additional analyses further indicate that the learned representations preserve action-sensitive and physically meaningful transition structure. These results suggest that supervising latent differences is an effective principle for learning compact, collapse-resistant world models for planning.

\bibliography{aaai2026}

\onecolumn
\appendix
\setcounter{secnumdepth}{2}
\section{Additional Probe and Attention Results}
\label{app:probe_attention}
\suppressfloats[t]

This appendix provides the complete diagnostic results that complement the probing and attention analyses in the main text. It is organized into three parts. Appendix~\ref{app:physical_state_probing} reports physical state probing results for Push-T, DMC Reacher, and OGB-Cube. Appendix~\ref{app:state_delta_probing} reports state-delta probing results on the same environments. Appendix~\ref{app:attention_specialization} provides an additional attention visualization showing layer-wise specialization in the visual encoder.

\subsection{Physical State Probing}
\label{app:physical_state_probing}

This section extends the physical state probing analysis beyond the Two-Room results reported in the main text. For each environment, we freeze the visual encoder and train linear and MLP probes to predict ground-truth physical quantities from latent states. Tables~\ref{tab:physical_probe_pusht}--\ref{tab:physical_probe_ogb_cube} report the results for Push-T, DMC Reacher, and OGB-Cube, covering controllable agent states, robot states, and object states. Lower MSE and higher Pearson correlation $r$ indicate that the learned representation preserves more physical information.

\begin{table}[t]
\centering
\caption{Physical latent probing results on Push-T. Lower MSE and higher $r$ indicate better representational quality.}
\label{tab:physical_probe_pusht}
\begin{tabular}{llcccc}
\toprule
& & \multicolumn{2}{c}{\textbf{Linear}} & \multicolumn{2}{c}{\textbf{MLP}} \\
\cmidrule(lr){3-4} \cmidrule(lr){5-6}
\textbf{Property} & \textbf{Method} & \textbf{MSE $\downarrow$} & \textbf{$r \uparrow$} & \textbf{MSE $\downarrow$} & \textbf{$r \uparrow$} \\
\midrule
\multirow{5}{*}{Agent Location}
& PLDM     & $0.007$ & $0.996$ & $0.000$ & $1.000$ \\
& Sub-JEPA & $0.094$ & $0.955$ & $0.003$ & $0.999$ \\
& LeWM     & $0.017$ & $0.991$ & $0.000$ & $1.000$ \\
& \textbf{Delta-JEPA} & $0.004$ & $0.998$ & $0.000$ & $1.000$ \\
\midrule
\multirow{5}{*}{Block Location}
& PLDM     & $0.055$ & $0.974$ & $0.006$ & $0.997$ \\
& Sub-JEPA & $0.250$ & $0.895$ & $0.006$ & $0.997$ \\
& LeWM     & $0.041$ & $0.979$ & $0.002$ & $0.999$ \\
& \textbf{Delta-JEPA} & $0.189$ & $0.929$ & $0.013$ & $0.994$ \\
\midrule
\multirow{5}{*}{Block Angle}
& PLDM     & $0.005$ & $0.998$ & $0.000$ & $1.000$ \\
& Sub-JEPA & $0.024$ & $0.988$ & $0.001$ & $0.999$ \\
& LeWM     & $0.004$ & $0.998$ & $0.000$ & $1.000$ \\
& \textbf{Delta-JEPA} & $0.011$ & $0.995$ & $0.001$ & $1.000$ \\
\bottomrule
\end{tabular}
\end{table}

\begin{table}[t]
\centering
\caption{Physical latent probing results on DMC Reacher. Lower MSE and higher $r$ indicate better representational quality.}
\label{tab:physical_probe_dmc_reacher}
\begin{tabular}{llcccc}
\toprule
& & \multicolumn{2}{c}{\textbf{Linear}} & \multicolumn{2}{c}{\textbf{MLP}} \\
\cmidrule(lr){3-4} \cmidrule(lr){5-6}
\textbf{Property} & \textbf{Method} & \textbf{MSE $\downarrow$} & \textbf{$r \uparrow$} & \textbf{MSE $\downarrow$} & \textbf{$r \uparrow$} \\
\midrule
\multirow{5}{*}{Finger Position}
& PLDM     & $0.016$ & $0.992$ & $0.000$ & $1.000$ \\
& Sub-JEPA & $0.798$ & $0.632$ & $0.014$ & $0.995$ \\
& LeWM     & $0.262$ & $0.869$ & $0.096$ & $0.969$ \\
& \textbf{Delta-JEPA} & $0.520$ & $0.777$ & $0.056$ & $0.972$ \\
\midrule
\multirow{5}{*}{Joint Position}
& PLDM     & $0.215$ & $0.879$ & $0.133$ & $0.928$ \\
& Sub-JEPA & $0.760$ & $0.446$ & $0.545$ & $0.673$ \\
& LeWM     & $0.630$ & $0.586$ & $0.789$ & $0.576$ \\
& \textbf{Delta-JEPA} & $0.622$ & $0.537$ & $0.555$ & $0.651$ \\
\bottomrule
\end{tabular}
\end{table}

\begin{table}[t]
\centering
\caption{Physical latent probing results on OGB-Cube. Lower MSE and higher $r$ indicate better representational quality.}
\label{tab:physical_probe_ogb_cube}
\begin{tabular}{llcccc}
\toprule
& & \multicolumn{2}{c}{\textbf{Linear}} & \multicolumn{2}{c}{\textbf{MLP}} \\
\cmidrule(lr){3-4} \cmidrule(lr){5-6}
\textbf{Property} & \textbf{Method} & \textbf{MSE $\downarrow$} & \textbf{$r \uparrow$} & \textbf{MSE $\downarrow$} & \textbf{$r \uparrow$} \\
\midrule
\multirow{5}{*}{Joint Position}
& PLDM     & $0.545$ & $0.595$ & $0.304$ & $0.813$ \\
& Sub-JEPA & $0.619$ & $0.482$ & $0.928$ & $0.511$ \\
& LeWM     & $0.817$ & $0.470$ & $1.494$ & $0.518$ \\
& \textbf{Delta-JEPA} & $0.378$ & $0.674$ & $0.621$ & $0.652$ \\
\midrule
\multirow{5}{*}{Joint Velocity}
& PLDM     & $0.953$ & $0.269$ & $0.656$ & $0.595$ \\
& Sub-JEPA & $1.082$ & $0.041$ & $1.797$ & $0.035$ \\
& LeWM     & $1.262$ & $0.054$ & $4.765$ & $0.035$ \\
& \textbf{Delta-JEPA} & $0.936$ & $0.273$ & $1.304$ & $0.283$ \\
\midrule
\multirow{5}{*}{End-Effector Position}
& PLDM     & $0.025$ & $0.988$ & $0.003$ & $0.998$ \\
& Sub-JEPA & $0.226$ & $0.909$ & $0.027$ & $0.987$ \\
& LeWM     & $0.515$ & $0.739$ & $0.256$ & $0.897$ \\
& \textbf{Delta-JEPA} & $0.007$ & $0.997$ & $0.001$ & $1.000$ \\
\midrule
\multirow{5}{*}{End-Effector Yaw}
& PLDM     & $0.363$ & $0.791$ & $0.136$ & $0.927$ \\
& Sub-JEPA & $0.958$ & $0.075$ & $1.874$ & $-0.084$ \\
& LeWM     & $1.468$ & $-0.037$ & $1.886$ & $-0.100$ \\
& \textbf{Delta-JEPA} & $1.218$ & $-0.074$ & $1.987$ & $-0.073$ \\
\midrule
\multirow{5}{*}{Block Position}
& PLDM     & $0.246$ & $0.860$ & $0.057$ & $0.971$ \\
& Sub-JEPA & $0.327$ & $0.835$ & $0.054$ & $0.973$ \\
& LeWM     & $0.464$ & $0.765$ & $0.244$ & $0.885$ \\
& \textbf{Delta-JEPA} & $0.038$ & $0.983$ & $0.007$ & $0.997$ \\
\midrule
\multirow{5}{*}{Block Quaternion}
& PLDM     & $0.635$ & $0.577$ & $0.296$ & $0.839$ \\
& Sub-JEPA & $1.180$ & $-0.030$ & $2.412$ & $-0.062$ \\
& LeWM     & $1.803$ & $-0.090$ & $3.670$ & $-0.057$ \\
& \textbf{Delta-JEPA} & $1.053$ & $0.273$ & $1.619$ & $0.205$ \\
\midrule
\multirow{5}{*}{Block Yaw}
& PLDM     & $0.462$ & $0.742$ & $0.190$ & $0.904$ \\
& Sub-JEPA & $1.696$ & $0.166$ & $3.156$ & $0.038$ \\
& LeWM     & $2.927$ & $-0.294$ & $2.945$ & $-0.076$ \\
& \textbf{Delta-JEPA} & $1.758$ & $0.241$ & $1.896$ & $0.232$ \\
\bottomrule
\end{tabular}
\end{table}

\subsection{State-Delta Probing}
\label{app:state_delta_probing}

This section evaluates whether latent displacements encode physical changes between consecutive observations. Instead of predicting state variables $x_t$ from $z_t$, each probe predicts $\Delta x_t=x_{t+1}-x_t$ from $\Delta z_t=z_{t+1}-z_t$. Tables~\ref{tab:state_delta_probe_pusht}--\ref{tab:state_delta_probe_ogb_cube} report the results for Push-T, DMC Reacher, and OGB-Cube, spanning agent motion, robot motion, end-effector motion, and object motion. This directly tests whether the transition representation preserves the direction and magnitude of environment changes.

\begin{table}[t]
\centering
\caption{State-delta probing results on Push-T. The probe predicts $\Delta x_t=x_{t+1}-x_t$ from $\Delta z_t=z_{t+1}-z_t$. Lower MSE and higher $r$ indicate better alignment between latent displacements and physical state changes.}
\label{tab:state_delta_probe_pusht}
\begin{tabular}{llcccc}
\toprule
& & \multicolumn{2}{c}{\textbf{Linear}} & \multicolumn{2}{c}{\textbf{MLP}} \\
\cmidrule(lr){3-4} \cmidrule(lr){5-6}
\textbf{Property} & \textbf{Method} & \textbf{MSE $\downarrow$} & \textbf{$r \uparrow$} & \textbf{MSE $\downarrow$} & \textbf{$r \uparrow$} \\
\midrule
\multirow{5}{*}{$\Delta$ Agent Location}
& PLDM     & $0.061$ & $0.969$ & $0.012$ & $0.994$ \\
& Sub-JEPA & $0.291$ & $0.846$ & $0.037$ & $0.981$ \\
& LeWM     & $0.091$ & $0.954$ & $0.017$ & $0.991$ \\
& \textbf{Delta-JEPA} & $0.018$ & $0.995$ & $0.001$ & $1.000$ \\
\midrule
\multirow{5}{*}{$\Delta$ Block Location}
& PLDM     & $0.114$ & $0.944$ & $0.020$ & $0.991$ \\
& Sub-JEPA & $0.375$ & $0.807$ & $0.028$ & $0.987$ \\
& LeWM     & $0.073$ & $0.966$ & $0.014$ & $0.993$ \\
& \textbf{Delta-JEPA} & $0.189$ & $0.927$ & $0.024$ & $0.989$ \\
\midrule
\multirow{5}{*}{$\Delta$ Block Angle}
& PLDM     & $0.103$ & $0.950$ & $0.012$ & $0.994$ \\
& Sub-JEPA & $0.333$ & $0.823$ & $0.015$ & $0.993$ \\
& LeWM     & $0.088$ & $0.959$ & $0.006$ & $0.997$ \\
& \textbf{Delta-JEPA} & $0.163$ & $0.924$ & $0.013$ & $0.994$ \\
\bottomrule
\end{tabular}
\end{table}

\begin{table}[t]
\centering
\caption{State-delta probing results on DMC Reacher. The probe predicts $\Delta x_t=x_{t+1}-x_t$ from $\Delta z_t=z_{t+1}-z_t$. Lower MSE and higher $r$ indicate better alignment between latent displacements and physical state changes.}
\label{tab:state_delta_probe_dmc_reacher}
\begin{tabular}{llcccc}
\toprule
& & \multicolumn{2}{c}{\textbf{Linear}} & \multicolumn{2}{c}{\textbf{MLP}} \\
\cmidrule(lr){3-4} \cmidrule(lr){5-6}
\textbf{Property} & \textbf{Method} & \textbf{MSE $\downarrow$} & \textbf{$r \uparrow$} & \textbf{MSE $\downarrow$} & \textbf{$r \uparrow$} \\
\midrule
\multirow{5}{*}{$\Delta$ Finger Position}
& PLDM     & $0.605$ & $0.654$ & $0.319$ & $0.828$ \\
& Sub-JEPA & $1.007$ & $0.197$ & $0.457$ & $0.751$ \\
& LeWM     & $0.784$ & $0.507$ & $0.644$ & $0.648$ \\
& \textbf{Delta-JEPA} & $0.900$ & $0.359$ & $1.150$ & $0.467$ \\
\midrule
\multirow{5}{*}{$\Delta$ Joint Position}
& PLDM     & $1.366$ & $0.067$ & $1.378$ & $0.284$ \\
& Sub-JEPA & $1.072$ & $-0.027$ & $1.065$ & $0.311$ \\
& LeWM     & $1.408$ & $0.017$ & $1.755$ & $0.137$ \\
& \textbf{Delta-JEPA} & $1.012$ & $0.207$ & $0.237$ & $0.870$ \\
\bottomrule
\end{tabular}
\end{table}

\begin{table}[t]
\centering
\caption{State-delta probing results on OGB-Cube. The probe predicts $\Delta x_t=x_{t+1}-x_t$ from $\Delta z_t=z_{t+1}-z_t$. Lower MSE and higher $r$ indicate better alignment between latent displacements and physical state changes.}
\label{tab:state_delta_probe_ogb_cube}
\begin{tabular}{llcccc}
\toprule
& & \multicolumn{2}{c}{\textbf{Linear}} & \multicolumn{2}{c}{\textbf{MLP}} \\
\cmidrule(lr){3-4} \cmidrule(lr){5-6}
\textbf{Property} & \textbf{Method} & \textbf{MSE $\downarrow$} & \textbf{$r \uparrow$} & \textbf{MSE $\downarrow$} & \textbf{$r \uparrow$} \\
\midrule
\multirow{5}{*}{$\Delta$ Joint Position}
& PLDM     & $0.851$ & $0.283$ & $2.056$ & $0.262$ \\
& Sub-JEPA & $0.744$ & $0.365$ & $0.613$ & $0.544$ \\
& LeWM     & $0.790$ & $0.355$ & $0.890$ & $0.517$ \\
& \textbf{Delta-JEPA} & $0.358$ & $0.686$ & $0.359$ & $0.711$ \\
\midrule
\multirow{5}{*}{$\Delta$ Joint Velocity}
& PLDM     & $1.188$ & $0.017$ & $2.478$ & $0.149$ \\
& Sub-JEPA & $1.076$ & $0.066$ & $1.056$ & $0.274$ \\
& LeWM     & $1.130$ & $0.055$ & $1.687$ & $0.253$ \\
& \textbf{Delta-JEPA} & $0.858$ & $0.391$ & $0.753$ & $0.555$ \\
\midrule
\multirow{5}{*}{$\Delta$ End-Effector Position}
& PLDM     & $0.608$ & $0.654$ & $0.261$ & $0.868$ \\
& Sub-JEPA & $0.443$ & $0.760$ & $0.113$ & $0.939$ \\
& LeWM     & $0.678$ & $0.568$ & $0.336$ & $0.812$ \\
& \textbf{Delta-JEPA} & $0.010$ & $0.995$ & $0.003$ & $0.999$ \\
\midrule
\multirow{5}{*}{$\Delta$ End-Effector Yaw}
& PLDM     & $1.147$ & $-0.012$ & $2.573$ & $-0.039$ \\
& Sub-JEPA & $0.851$ & $0.083$ & $1.203$ & $0.188$ \\
& LeWM     & $1.055$ & $0.037$ & $1.578$ & $-0.013$ \\
& \textbf{Delta-JEPA} & $0.851$ & $0.239$ & $0.160$ & $0.910$ \\
\midrule
\multirow{5}{*}{$\Delta$ Block Position}
& PLDM     & $0.845$ & $0.417$ & $1.047$ & $0.420$ \\
& Sub-JEPA & $0.614$ & $0.590$ & $0.337$ & $0.790$ \\
& LeWM     & $0.675$ & $0.526$ & $0.539$ & $0.646$ \\
& \textbf{Delta-JEPA} & $0.198$ & $0.886$ & $0.024$ & $0.987$ \\
\midrule
\multirow{5}{*}{$\Delta$ Block Quaternion}
& PLDM     & $0.927$ & $0.027$ & $1.921$ & $0.087$ \\
& Sub-JEPA & $0.827$ & $-0.004$ & $1.290$ & $0.028$ \\
& LeWM     & $0.911$ & $0.023$ & $1.273$ & $0.108$ \\
& \textbf{Delta-JEPA} & $0.688$ & $0.226$ & $1.001$ & $0.159$ \\
\midrule
\multirow{5}{*}{$\Delta$ Block Yaw}
& PLDM     & $1.668$ & $-0.055$ & $34.418$ & $-0.085$ \\
& Sub-JEPA & $1.562$ & $-0.012$ & $1.651$ & $-0.025$ \\
& LeWM     & $1.638$ & $-0.041$ & $3.424$ & $-0.013$ \\
& \textbf{Delta-JEPA} & $1.574$ & $0.050$ & $1.592$ & $0.043$ \\
\bottomrule
\end{tabular}
\end{table}
\subsection{Layer-Wise Attention Specialization}
\label{app:attention_specialization}

This section complements the attention rollout visualizations in the main text by examining whether different encoder layers emphasize different task-relevant entities. We visualize OGB-Cube attention maps from two intermediate layers of the same encoder to compare how attention shifts across the visual hierarchy.

\begin{figure}[t]
\centering
\includegraphics[width=\textwidth]{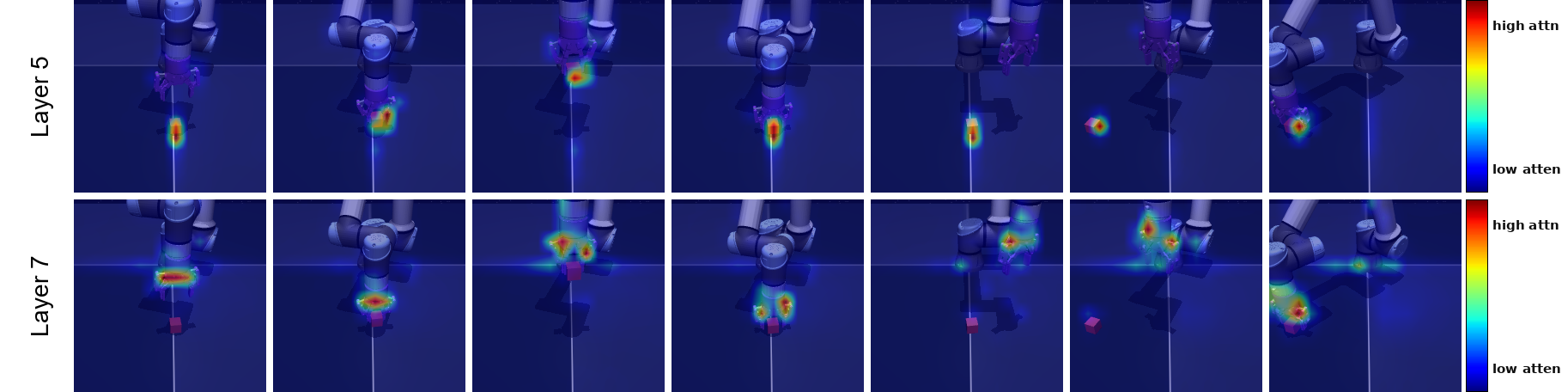}
\caption{Layer-wise specialization of attention maps on OGB-Cube. Layer 5 highlights the target cube, while layer 7 more prominently attends to the robotic gripper. Warmer colors indicate higher attention weights.}
\label{fig:layer_specialization}
\end{figure}

As illustrated in Figure~\ref{fig:layer_specialization}, different intermediate layers emphasize distinct functional components of the same OGB-Cube scenes. Layer 5 primarily attends to the target cube, whereas layer 7 places stronger emphasis on the robotic gripper. These observations indicate that the encoder progressively organizes task-relevant entities across layers, rather than relying on a single undifferentiated saliency pattern.

\end{document}